\documentclass{article}

\usepackage{microtype}
\usepackage{graphicx}
\usepackage{subfigure}
\usepackage{amsmath}
\usepackage{booktabs} 
\usepackage{authblk}

\usepackage{enumitem}

\usepackage{hyperref}

\allowdisplaybreaks

\usepackage{todonotes}

\newcommand{\dlcontrib}{C_{\Delta x \Delta t}}

\usepackage[accepted]{icml2020}

\icmltitlerunning{Deep interpretability for GWAS}

\begin{document}

\twocolumn[
\icmltitle{Deep interpretability for GWAS}




\begin{icmlauthorlist}
\icmlauthor{Deepak Sharma}{mcg,mila}
\icmlauthor{Audrey Durand}{laval,mila}
\icmlauthor{Marc-André Legault}{udem}
\icmlauthor{Louis-Philippe Lemieux Perreault}{mhi,bspc}
\icmlauthor{Audrey Lemaçon}{udem,mhi,bspc}
\icmlauthor{Marie-Pierre Dubé}{udem,mhi,bspc}
\icmlauthor{Joelle Pineau}{mcg,mila,fair}
\end{icmlauthorlist}

\icmlaffiliation{mcg}{Computer Science, McGill University, Montreal, Quebec, Canada}
\icmlaffiliation{mila}{MILA - Quebec AI Institute, Montreal, Quebec, Canada}
\icmlaffiliation{laval}{Université Laval}
\icmlaffiliation{udem}{Faculty Of Medicine, Université de Montréal, Montreal, Quebec, Canada}
\icmlaffiliation{mhi}{Montreal Heart Institute, Montreal, Quebec, Canada}
\icmlaffiliation{bspc}{Beaulieu-Saucier Pharmacogenomics Centre, Montreal, Quebec, Canada}
\icmlaffiliation{fair}{Facebook AI Research, Montreal, Quebec, Canada}

\icmlcorrespondingauthor{Deepak Sharma}{deepak.sharma@mail.mcgill.ca}


\vskip 0.3in
]




\printAffiliationsAndNotice{} 

\begin{abstract}
Genome-Wide Association Studies are typically conducted using linear models to find genetic variants associated with common diseases. In these studies, association testing is done on a variant-by-variant basis, possibly missing out on non-linear interaction effects between variants. Deep networks can be used to model these interactions, but they are difficult to train and interpret on large genetic datasets. We propose a method that uses the gradient based deep interpretability technique named DeepLIFT to show that known diabetes genetic risk factors can be identified using deep models along with possibly novel associations.

\end{abstract}

\section{Introduction}
\label{introduction}

A Genome-Wide Association Study (GWAS) ~\cite{linearmodelsgwas} aims at identifying common genetic variants associated with a trait of interest. Various diseases, e.g. diabetes, and continuous measurements, e.g. LDL-cholesterol, have been studied using GWAS.  They are typically conducted using Single-Nucleotide Polymorphisms (SNPs) scattered across the genome, which are genetic variants consisting of a single base-pair change in DNA. The SNPs are tested one-by-one using multifield linear models relating them to the target outcome, and statistical hypothesis testing is used to find variants with non-null effects. 

Even though GWAS associations can't directly be interpreted as causal, they can help predict risk of complex diseases, identify possible drug targets or gain insight into disease biology by highlighting relevant genetic loci.
Despite their many successes, GWAS are limited to the identification of genetic variants with strong marginal effects, possibly leaving out a large number of genetic effects governed by SNP-SNP interactions or other non-linear effects. These  effects may play an important role in complex diseases ~\cite{diabetes}. Since deep networks are known to be able to model arbitrarily complicated non-linear functions of their inputs~\cite{Goodfellow-et-al-2016}, we aim to use them to predict complex disease outcomes from genetic data, and then interpret them (sample by sample) to discover novel interactions between SNPs that are significant predictors of the disease outcome.

Previous works~\cite{tran2018implicit,Wang2019} have formulated GWAS as a problem of causal inference with multiple causes of (typically) a single trait and with hidden confounding. They use deep networks to model complex traits on both simulated as well as real datasets with up to approximately 300,000 SNPs and 5,000 individuals. They proceed to conduct a likelihood ratio test on the fitted deep models to identify strongly associated SNPs. Likelihood ratio testing for each SNP can be resource and time consuming, when considering hundreds of thousands of SNPs, as it requires training two distinct models per SNP.

In this work, we use DeepLIFT~\cite{deeplift}, a reference-based feature attribution technique, to identify the SNPs used by a feedforward model trained to identify individuals with a given trait. This allows us to train and interpret a single deep model to yield a list of potentially causal SNPs. We use the implementation found in Pytorch-Captum ~\cite{captum} to interpret the trained models. \citet{MoovaDeepLift} use DeepLift on a Convolutional Neural Network (CNN) to identify SNPs that could potentially regulate gene activity. They use the predictions of their model to supplement GWAS results and isolate causal variants. In contrast, we are using DeepLIFT to find putative causal variants by directly predicting traits of interest.

We first summarize the proposed methods for conducting a GWAS using a deep learning pipeline (Sec.~\ref{sec:methods}). We then consider simulated (Sec.~\ref{sec:data:sim}) as well as real (Sec.~\ref{sec:data:ukbb}) data to conduct a set of experiments where we show that we are able to identify the SNPs for predicting discrete and continuous traits (Sec.~\ref{sec:xps_res}). \footnote{The code to reproduce the simulation results can be found \href{https://drive.google.com/file/d/1kP9gZtNjBCkldGAywW6mcPyDfeKNBRzx/view?usp=sharing}{here} or (shorturl.at/oT056). This research has been conducted using the UK Biobank Resource under Application Number 20168.}

\vspace{-2mm}
\paragraph{Contributions}

We believe that this work is the first demonstration of the applicability of recent progress in interpretability of deep models to GWAS. We have shown that there is a considerable overlap between DeepLIFT identified variants and known GWAS signals. We provide a working pipeline with both simulated and real data. This lays the ground to the next step, which is to further characterize associations identified by DeepLIFT and not by conventional GWAS and to relate them to SNP-SNP interaction effects.

\section{Methods}
\label{sec:methods}


Consider a dataset $\mathcal D$ of $N$ individuals with $M$ SNPs.

\vspace{-3mm}
\paragraph{Step 1: Train a feedforward model}

Using a training set $\mathcal D_T \subset \mathcal D$, we train a feedforward model to output a prediction $\hat t_n$ given input SNPs $z_n$. One can use the typical Mean Squared Error or Cross Entropy as a loss for that task.

\vspace{-3mm}
\paragraph{Step 2: Apply DeepLIFT}

Using a validation set $\mathcal D_V \subset \mathcal D$ such that $\mathcal D_V \cap \mathcal D_T = \emptyset$, we use DeepLIFT~\cite{deeplift} to identify the saliency of input points according to how sensitive the output of the network is to the presence of a certain input feature compared to its \textit{baseline} (or reference value). This reference value is picked appropriately for the question at hand. In GWAS, the question is whether mutations of SNPs from their expected values plays a role in a disease. Therefore, we pick the mean allele frequency for each SNP as our reference value. This allows to visualize the saliency of SNPs in the dataset against their average.

More precisely, let $x$ denote the input neurons for our model and let $t$ be the target output neuron. Let $x^{\text{ref}}$ and $t^{\text{ref}}$ be the input and target neuron outputs when the model receives the reference input. Let $\Delta x = x - x^{\text{ref}}$ and $\Delta t = t - t^{\text{ref}}$ be  the difference in the outputs of the input and target neurons from their respective reference outputs.

DeepLIFT assigns a score for each input neuron $x_i$ s.t.
\vspace{-2mm}
\begin{equation}
    \label{eq:deeplift_relevance}
    \Delta t = \sum_{i=1}^{L} C_{\Delta x_i \Delta t},
\end{equation}
where $L$ is the number of input neurons and $C_{\Delta x_i \Delta t}$ can be thought of as a weight assigned to each input neuron in proportion to its contribution to the difference $\Delta t$.

In order to compute the contribution score $\dlcontrib$, DeepLIFT defines the multiplier $m_{\Delta x \Delta t}$ as
\vspace{-2mm}
\begin{equation}
    \label{dlmultiplier1}
    m_{\Delta x \Delta t} = \frac{\dlcontrib}{\Delta x},
\end{equation}
and a Chain Rule to propagate the multiplier values from output neurons to the input neurons:
\vspace{-2mm}
\begin{equation}
    \label{dlmultiplier2}
    m_{\Delta x_i \Delta t} = \sum_{j=1}^{L} m_{\Delta x_i \Delta y_j} m_{\Delta y_j \Delta t},
\end{equation}
where $y_j$ is a neuron that is immediately downstream to the neuron $x_i$ and $L$ is the number of neurons in the layer immediately downstream to $x_i$.

We use the implementation of DeepLIFT found in Pytorch-Captum~\cite{captum}, which references~\citet{ancona2018towards} for assigning an attribution score $A_i$ to each input feature $i$ from the contribution scores of each input neuron $C_{\Delta x \Delta t}$, using the RESCALE rule~\cite{deeplift}.


\vspace{-3mm}
\paragraph{Step 3: Identify \textit{impactful} SNPs}

With DeepLIFT, positive and negative attribution scores indicate that a feature contributed to bringing the value of the target up or down respectively, relative to the reference target. The magnitude of the attribution score reflects how strongly a model relied on that feature for a particular prediction. The most impactful SNPs will consistently have the highest magnitudes. Therefore, we take the mean of the absolute value of the attribution scores for each input feature.

\vspace{-3mm}
\paragraph{Robustness and reproducibility}

Given that genetic datasets contain complex correlations between causal and non-causal SNPs, it is important to verify the reproducibility and robustness of the attribution scores by repeating the training and model interpretation on multiple seeds. This is addressed in our experiments (Sec.~\ref{sec:xps_res}).


\section{Data}

We conduct experiments (Sec.~\ref{sec:xps_res}) on the following datasets.

\subsection{Simulated}
\label{sec:data:sim}

We follow the procedure in \citet{HaoSpatial} to simulate $10,000$ SNPs, $10,000$ samples, and their corresponding binary traits. In short, the $M \times N$ genotype matrix is sampled from a Binomial distribution: $x_{mn} \sim \operatorname{Binomial}(2, \pi_{mn})$ with $\pi = \Gamma S$, where $\Gamma$ is a $M \times 3$ matrix and $\Gamma_{mk} \sim \operatorname{Uniform}(0, 0.5)$ for $k=1,2$ and $\Gamma_{m3} = 0.5$. The $3 \times N$ matrix $S$ has $S_{kn} \sim \operatorname{Beta}(a,a)$ for $k = 1,2$ and $S_{3n} = 1$. The first 2 rows of S correspond to the position of each sample on a unit square. Smaller values of the $\mathcal B$ parameter $a$ clusters the population into the four corners of a unit square.

The traits are sampled from Bernoulli distributions with parameters that are a function of a random effect vector, the SNPs, and the spatial position of each sample:
\vspace{-2mm}
\begin{align*}
    y_n & \sim \mathcal B \left(\sigma(\sum_{m=1}^{M} \beta_m x_{mn} + \lambda_n + \epsilon_n) \right) \\
    \epsilon_n & \sim \mathcal N(0, \sigma_n^2) \\
    \beta_m & \sim \begin{cases}
      \mathcal N(0, 1), & \text{if}\ m \leq 10 \\
      0, & \text{if}\ m > 10.
    \end{cases}
\end{align*}
As per \citet{HaoSpatial} and \citet{tran2018implicit}, we simulate $\lambda_n$ and $\sigma_n$ as follows:
\begin{enumerate}[topsep=0pt,itemsep=-1ex,partopsep=1ex,parsep=1ex]
    \item Assign each sample $j$ to a partition obtained by running $K$-mean clustering on the columns of the sample frequency matrix $S$, with $K = 3$. Let the partitions be denoted by $S_1$, $S_2$, and $S_3$
    \item $\lambda_j = k$ for all $j \in S_k$
    \item Draw $\tau_1^2, \tau_2^2, \tau_3^2 \sim \operatorname{InverseGamma}(3, 1)$ and set $\sigma_j^2 = \tau_k^2,$ $\forall j \in S_k$.
\end{enumerate}

We produce 20 different simulated datasets using four different values of $a \in \{ 0.01, 0.1, 0.5, 1 \}$ over five seeds.



\subsection{UK Biobank}
\label{sec:data:ukbb}


The UK Biobank is a population cohort including more than 500,000 genotyped participants
\cite{ukbb}.
%
Using the following pre-processing, we build two datasets for the phenotypes diabetes and glycated hemoglobin measurements (HbA1c).

\vspace{-3mm}
\paragraph{Filtering individuals}

We consider the imputed genotypes, where we filter out individuals with more than 2\% missing genotypes, in addition to individuals with sexual chromosome aneuploidies or with genetically inferred sex different from the self-reported sex. We then select a subset of the cohort of European ancestry to avoid population stratification (\textit{i.e.} confounding bias due to ethnicity) by using the UK Biobank provided principal components and keeping individuals near the cluster of individuals self-reporting as of \textit{white British ancestry}. To avoid including related individuals, we randomly select one individual from pairs with a kinship coefficient above 0.0884 (corresponding to a 2nd degree relationship). This results in the selection of $N=413,173$ individuals (samples).

\vspace{-3mm}
\paragraph{Filtering variants}

After selecting the individuals to include, we filter genetic variants to be used as features. Starting from all variants on chromosome 10, we filter out variants with minor allele frequency under 1\%, variants with a call rate under 99\% and set genotypes with a probability under 90\% to missing. This results in the selection of $M=336,814$ variants (SNPs) per individual.

\vspace{-3mm}
\paragraph{Phenotype extraction} 
The Diabetes phenotype is defined based on a combination of hospitalization codes and the self-reported verbal interview data. Specifically, we code as cases any participant with data coding for Diabetes (field \#20002, coded as 1220)
as cases, or with the `249' or `250' ICD9 codes, or E10, E11, E12, E13, or E14 ICD10 codes as the primary or secondary reason for hospitalization. The remaining individuals are used as controls. This results in a dataset with 24,717 diabetes cases and 388,456 controls.

The HbA1c phenotype (field \#30750) is extracted for 395,042 participants. If multiple measurements are available for an individual, the arithmetic mean is used. The values are also log-transformed to ensure an approximately normal distribution as typical in continuous trait GWAS.

\section{Experiments and Results}
\label{sec:xps_res}

We now highlight the potential of the proposed methods.


\subsection{Predicting a discrete simulated trait}
\label{sec:xps:simulated_discrete}

We first conduct a simulation study (Sec.~\ref{sec:data:sim}) to validate that deep network feature attribution techniques can be used to identify causal SNPs. We randomly generate 20 datasets of 10,000 individuals with 10,000 SNPs, including 10 causal SNPs ($a\in \{0.01,0.1,0.5,1\}$ for five seeds).
We train several two-layer (first in $\{64,128\}$, second in $\{128,256\}$) feedforward models to predict a simulated binary trait. We train each model on 50\% of the samples, early stop on 25\% of the samples, and validate the model architectures using the remaining 25\%. We force the models to pick a handful of SNPs by adding an L1-penalty ($\lambda\in\{0.01,0.1,1,10\}$) to the first layer weights.
We then use DeepLIFT to compute a summary score for each input SNP. We set the reference input to be the mean genotype of all samples.



\vspace{-3mm}
\paragraph{Model selection}

We pick the model with the best log likelihood on the validation set averaged over five seeds, per spatial configuration.
We verify the robustness of the performance of the selected model by re-training and re-running DeepLIFT on the same dataset with five more seeds. Thus for each of the four selected model architectures, we perform 25 trainings, on five different datasets with the same value of $a$, five times each. We consider the SNPs with the 10 highest DeepLIFT scores (in magnitude) as causal.

\vspace{-3mm}
\paragraph{Results}



\begin{table}[t]
    \centering
        \begin{tabular}{|r|l|l|r|}
                \toprule
                Popl. sparsity $a$ &     arch &  L1 coeff. & Recall\\
                \midrule
                0.01 &   64 by 256 &       1 & 64\% \\ 
                0.10 &  128 by 256 &       1 & 66\% \\
                0.50 &  128 by 256 &       1 & 70\% \\
                1.00 &   64 by 256 &       1 & 66\% \\
                \bottomrule
        \end{tabular}
    \caption{Models with the highest mean validation likelihood,  averaged over 5 seeds, for each simulation setting $a$. We are able to capture at least 64\% of the causal SNPs in all settings.}
    \label{tab:best_sim}
\end{table}


\begin{figure}[t]
\begin{center}
\centerline{\includegraphics[width=\columnwidth]{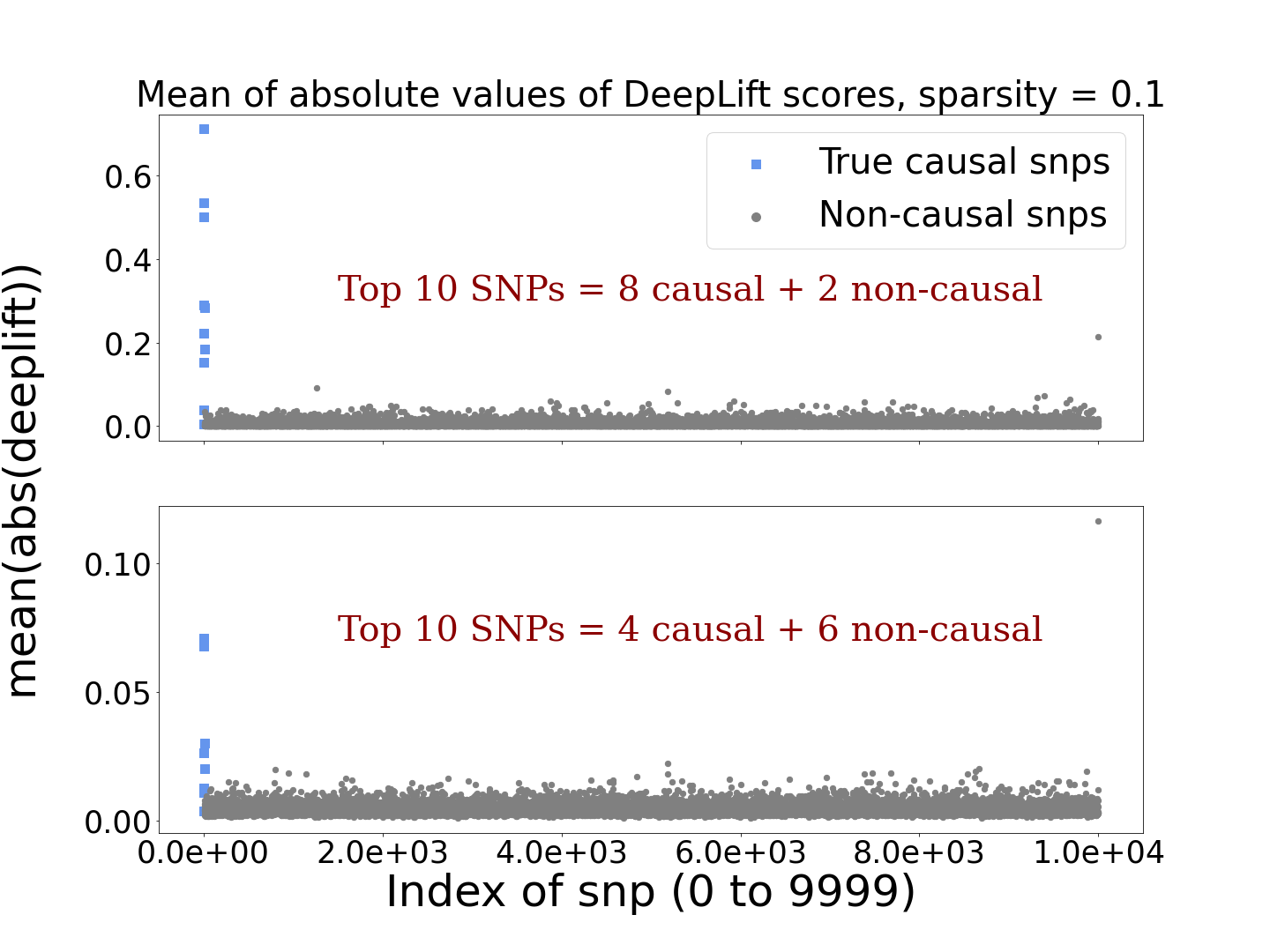}}
\vspace{-3mm}
\caption{Mean absolute DeepLIFT scores with the same architecture on two different seeds}
\label{fig:sim_res_01}
\end{center}
\end{figure}

Table~\ref{tab:best_sim} reports the architectures of the best performing models for each configuration, and the corresponding average Recall (number of true positives divided by total number of positives) over the 25 different runs.
Figure~\ref{fig:sim_res_01} shows two interpretability plots of a model trained and interpreted using two different seeds on the same dataset ($a = 0.1$). Both models have comparable negative log likelihood on the validation set (0.25 vs 0.31) but the number of identified causal SNPs is different. This shows how the same model can come to rely on different subsets of features on a dataset consisting of highly correlated features (as is typical in genetic datasets). This necessitates training and interpreting a model on multiple seeds to verify interpretability results.

\subsection{Predicting Diabetes and HbA1c}
\label{sec:xps:diabetes}

In order to test the scalability and robustness of our approach on real data, we perform experiments on genetic data obtained from the UK Biobank (Sec.~\ref{sec:data:ukbb}).
We train several two-layer (first in $\{32,64,128\}$, second in $\{64,128,256\}$) feedforward models to predict diabetes and hba1c, while again applying the L1-penalty ($\lambda\in\{0.01,0.1,1,10\}$).

We evaluate the proposed approach by comparing with GWAS analyses adjusted for age (field \#21022), sex (field \#31), and the first 10 ethnicity principal components as provided by the UK Biobank.


\vspace{-3mm}
\paragraph{Model selection}

We pick the model with the highest log likelihood on the validation set. We then re-train this model and conduct feature attribution using five random seeds.

\vspace{-3mm}
\paragraph{Results}

\begin{figure}[ht!]
\begin{subfigure}
  \centering
  \includegraphics[width=0.95\columnwidth]{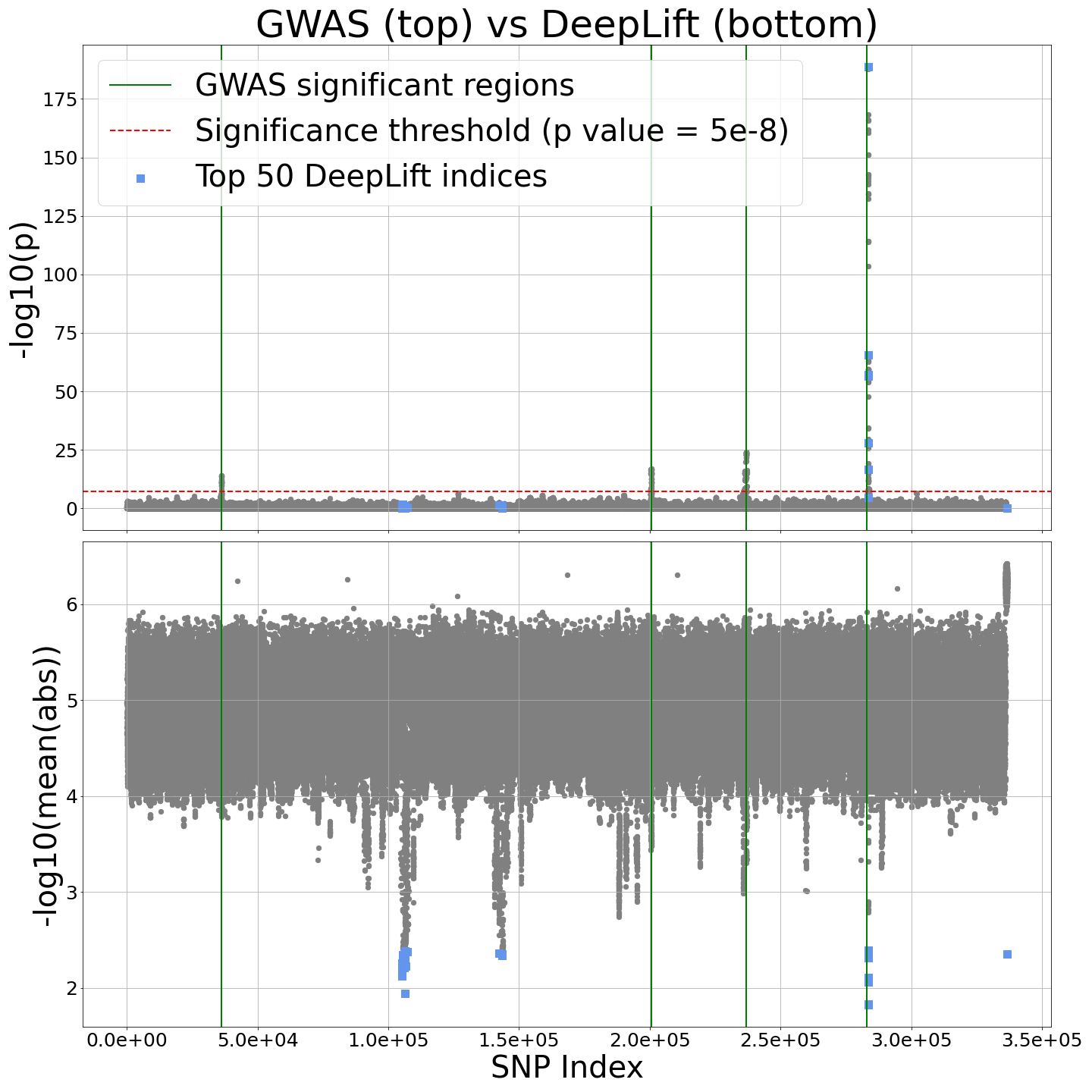}
\end{subfigure}
\vspace{-3mm}
\begin{subfigure}
  \centering
  \includegraphics[width=0.95\columnwidth]{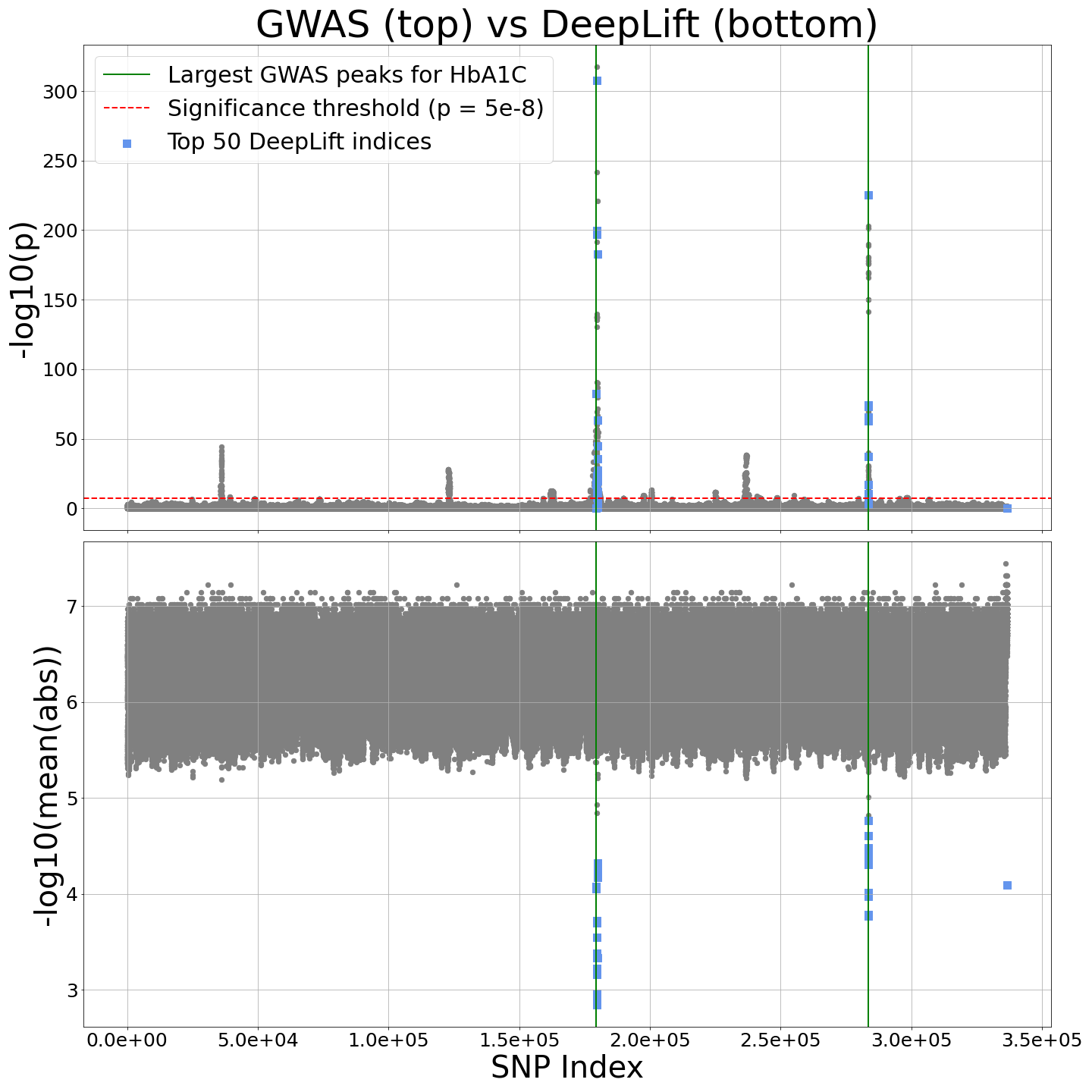}
\end{subfigure}
\caption{Miami plot of a linear GWAS against mean absolute DeepLIFT scores on Diabetes (top) and HbA1C (bottom). The best performing Diabates model had 64 by 256 hidden units, and the best HbA1C model had 128 by 256 units. The L1 regularization parameter was 1 for both models.}
\label{fig:ukbb_miami}
\vspace{-3mm}
\end{figure}

Fig.~\ref{fig:ukbb_miami} shows their corresponding Miami plots. We observe that for both traits, the topmost peaks in both halves of their respective Miami plots occur at the same index. There is also some overlap between the secondary peaks for Diabetes, as indicated by the second and third vertical lines from the right. For HbA1c, the two tallest peaks are well captured by both GWAS and DeepLIFT.

\vspace{-3mm}
\section{Conclusion}

We showed that deep neural networks trained on large genetic data can be interpreted to uncover interactions that a normal GWAS would not identify. This provides a way forward for deep models to be used for discovery of novel interactions in genomics. We are also excited at the prospect of capturing novel variants while predicting two or more traits together using multi-task learning.
Although the feature attribution techniques in deep networks are not as rigorously grounded as statistical testing of models, we hope that we were able to show that they can provide useful and reproducible results for exploratory analysis on real-world, large biological datasets.
In this work, we did not attempt to control for confounding in the deep learning model, or tackle the imbalance between cases and controls in the diabetes dataset. As future work, we will mitigate the former by including age, sex, and principal components provided by the UKBB, and the latter by oversampling the cases during training. This should help reduce the number of potentially spurious associations that will need to be validated.






\bibliography{example_paper}
\bibliographystyle{icml2020}
\end{document}